# Shapley values for cluster importance

## How clusters of the training data affect a prediction

Andreas Brandsæter[1,2] · Ingrid K. Glad[2]

**Abstract**
This paper proposes a novel approach to explain the predictions made by data-driven methods. Since such predictions rely heavily on the data used for training, explanations that convey information about how the training data affects the predictions are useful. The paper proposes a novel approach to quantify how different data-clusters of the training data affect a prediction. The quantification is based on Shapley values, a concept which originates from coalitional game theory, developed to fairly distribute the payout among a set of cooperating players. A player's Shapley value is a measure of that player's contribution. Shapley values are often used to quantify feature importance, ie. how features affect a prediction. This paper extends this to cluster importance, letting clusters of the training data act as players in a game where the predictions are the payouts. The novel methodology proposed in this paper lets us explore and investigate how different clusters of the training data affect the predictions made by any black-box model, allowing new aspects of the reasoning and inner workings of a prediction model to be conveyed to the users. The methodology is fundamentally different from existing explanation methods, providing insight which would not be available otherwise, and should complement existing explanation methods, including explanations based on feature importance.

---

Responsible editor: Martin Atzmueller, Johannes Fürnkranz, Tomáš Kliegr and Ute Schmid

---

✉ Andreas Brandsæter
andreas.brandsaeter@hivolda.no

Ingrid K. Glad
glad@math.uio.no

1 Department of Science and Mathematics, Volda University College, Volda, Norway

2 Department of Mathematics, University of Oslo, Oslo, Norway

## 1 Introduction

There is an increasing interest in and demand for interpretations and explanations of machine learning models and their predictions in various application areas (Rai 2020; Islam et al 2022). Their reasonings can sometimes be intentionally hidden from us, but most often they are unavailable due to the complexity of the systems and the models used. The algorithms can be simple enough, but after training on massive and complex datasets, the final models are often difficult to decipher and challenging to explain and interpret. Due to the models' inscrutable inner workings, such models are often labelled black-boxes (Hall and Gill 2018).

The importance of transparency, explanations and interpretations of machine learning models is growing, particularly for decision making in high risk and safety critical applications (Kim et al 2016), including for example clinical decision support systems (Antoniadi et al 2021) for example for cancer detection, distinguishing between fraudulent and genuine claims to an insurance company (Rawat et al 2021), autonomous navigation systems supervised by humans (Brandsæter et al 2020), or decision support systems in law enforcement intended to improve legal practice (Metsker et al 2021). Ribeiro et al (2016) claim that "if the users do not trust a model or a prediction, they will not use it". If we understand the model's reasoning, it is easier to verify the model and determine when the model's reasoning is in error, and to improve the model (Caruana et al 1999; Doshi-Velez and Kim 2017; Lundberg and Lee 2017). Furthermore, transparency, interpretations and explanations can help us guard against unethical or biased predictions, such as discriminations, and we can better deal with competing objective functions of the algorithms, such as privacy and prediction quality (Doshi-Velez and Kim 2017). Interpretation also lets us learn from the model, and convert interpretations and explanations into knowledge (Shrikumar et al 2016). Moreover, the EU General Data Protection Act (GDPR) provides individuals the right to receive an explanation for algorithmic decisions which significantly affect that individual (Goodman and Flaxman 2017).

But what is a good explanation? Lipton (2016) discusses the interpretability of human decision-makers, and what notion of interpretability these explanations satisfy. He argues that human explanations do not clarify the mechanisms or the precise algorithms by which brains work. Nevertheless, the information conferred by an interpretation may be useful. Hence, Doshi-Velez and Kim (2017) propose to define interpretability as "the ability to explain or to present in understandable terms to a human." When facing a problem, we can base our decision on previous experiences from facing similar problems. It can therefore be meaningful to refer to these previous experiences when explaining our decision. Suppose you face a problem in your new job, how does experience from your previous jobs affect your decision? Similarly, when interpreting the predictions of a machine learning model, it can be meaningful to quantify how different parts of the training data affect the prediction.

*Contribution:* In this paper, we propose a novel data-centric influence measure which we call Shapley values for cluster importance. The Shapley value concept originates from coalitional game theory, and it is well-established to quantify the importance of the different features (explanatory variables) of a prediction model using Shapley values by letting the features act as collaborating players in a game where the prediction is the payout. We adapt the calculation of Shapley values to cluster importance, letting clusters of the training data be the collaborating players. This allows users of a prediction method to quantify how different clusters in the training data affect individual predictions. This information can for example help lay users and experts to better understand limitations of the performance of the model, to reveal discriminatory behavior in its models, to investigate biases arising from different sources of data, as well as to reveal potential erroneous data. The proposed methodology is fundamentally different from existing explanation methods, and should complement existing explanation methods based on feature importance. The interest to study the importance of the clusters, is to quantify how different parts of the training data influence a specific prediction from the machine learning model in question. The division of the training data into clusters might be done in various ways, depending on the type of data. The clusters do not have to be discovered algorithmically, in the traditional sense of clustering, but can be manually defined by experts and so be based on any variable or combination of variables. When using the proposed Shapley values for cluster training data importance, it should always be kept in mind how the clusters of training data have been formed.

Existing methods and measures from influential statistics such as Cook's distance (Cook 1977, 1979) are already an essential part of best practice data analysis and model interpretation. Cook's distances let us identify individual data-points that are particularly influential, but the combined influence of several instances and their interactions are not available. This is problematic since interactions between the data-points can strongly influence model training and prediction (Molnar 2021, Ch. 6). We overcome this challenge, both for individual points and clusters by using an approximation method similar to the well-established method to approximate Shapley values for feature importance (See Seq. 2.3).

In the following, we first provide an overview of related work and available XAI and machine learning interpretation methods. We present the theoretical background for Shapley values, including its extension to feature importance. In Sect. 3, we describe our proposed novel metric: Shapley value for cluster importance, and explain how we can calculate and approximate it building on the method for calculating Shapley values for feature importance. In Sect. 4, we provide a set of illustrative examples sketching how the proposed measure can be used. Finally, we discuss future work, challenges and limitations in Sect. 5, and conclude in Sect. 6.

## 2 Background

### 2.1 Related interpretation methods

One way to achieve interpretability is to use interpretable models, such as linear regression, logistic regression and decision trees. However, one can argue that sufficiently high-dimensional models, for example deep decision trees, can be considered less transparent than comparatively compact neural networks. Several methods have been proposed and developed to interpret the black-box models and explain their predictions. Some of these methods are model-specific, that is, they can only be used on specific machine learning models, while other methods are model-agnostic, and these are the focus of this study. If a task should be solved with machine learning methods, typically, several types of models are evaluated. The use of model-agnostic explanation methods allows us to compare different models in terms of interpretability (Molnar 2021, Ch. 5).

Counterfactual explanations is an increasingly popular class of explanation methods. Such methods seek to explain a prediction by showing how a small change in the input feature would affect the output (Verma et al 2020). Such explanations can, however, be vulnerable to issues caused by lack of robustness of the classifier. Hence, Laugel et al (2019a, b) argue that such explanations should be justified, meaning that a counterfactual instance should be continuously connected to an observation from the training dataset. Counterfactual explanations are closely linked to adversarial examples and adversarial attacks where features are perturbed intentionally to cause a false prediction (Molnar 2021, Ch. 6).

Since the predictions made by the data-driven methods rely heavily on the training data used, we also advocate explanations which convey how the training data affects the predictions. This includes case-based explanation methods, which select particular observations of the dataset to explain the behavior of machine learning models. Caruana et al (1999) propose a method to generate case-based explanations for non-case-based learning methods, claiming it to be very useful especially in medical applications, since medical training and practice emphasize case evaluation. In general, case-based explanation methods work well if the feature values of a specific data point carry some context, meaning the data has a structure, like images or texts (Molnar 2021, Ch. 5).

Similarly, Koh and Liang (2017) suggest that we can better understand a model's behavior by studying how the model is derived from its training data, and propose to identify training points most responsible for a given prediction. For linear models and generalized linear models, the influence of specific data points in the training data are commonly estimated using Cook's distance (Cook 1977, 1979) or similar. Koh and Liang (2017) use influence functions which tell us how the model parameters change when a point in the training dataset is up-weighted by an infinitesimal amount. Approximations to these influence functions are claimed to provide valuable information even on non-convex and non-differentiable models where the theory breaks down.

The influence measures outlined above only take into account the influence of individual data points, disregarding interactions between them. For example, for some machine learning models, if two points in the training dataset are duplicates, removing one of them will not influence the model, while removing both will significantly change the model. For example, for a $k$-nearest neighbor model, if we have $k + l$ identical points, removing $l$ of them will not change the predictions. Unfortunately, if we try to systematically delete combinations of points from the training data, the number of possible combinations explodes. A quantification of the importance of each of the points in the full training data is also difficult to interpret due to the large size of the data.

Koh and Liang (2017) suggest that sometimes we might be interested in broader effects, rather than from individual observations, such as for for example how a sub-population of patients from a specific hospital affects a fitted model. They argue that since influence functions depend on the model not changing too much, how to analyze the effect and importance of subsets of the training data is an open problem. Hence, in this paper, we propose a model agnostic method to explain individual predictions by quantifying how different clusters of the training data affect the predictions. We propose to use Shapley values to approximate the importance of the different clusters, taking interactions between clusters into account. When Shapley values are used to calculate and estimate feature importance, the features act as players in a game where the predictions are the payouts. In our proposed methodology, the clusters replace the features as players. Hence, we call the new measure the Shapley value for cluster importance. Case-based explanations work well when the feature values carry context. Similarly, when the clusters carry context, and the training data can be divided into clusters based on some inherent structure, we believe our proposed explanations provide valuable information.

A frequently used model-agnostic approach to interpret and explain the decisions and predictions made by machine learning algorithms is the concept of feature importance. For a linear regression model, the importance of different features is readily available, and various methods aim to provide a similar interpretation of more complex models. A feature's relative importance can for example be estimated by perturbing the values of the test point, and observing and analysing how the prediction changes (Breiman 2001; Fisher et al 2018). Another approach is to approximate the black-box model with an interpretable surrogate model, and base the explanation on the surrogate. Ribeiro et al (2016) propose a local surrogate method, LIME, which approximates any machine learning model locally with an interpretable model (for example a linear model), and use this model to explain individual predictions.

Yet another popular estimate of local feature importance is the so called Shapley value. As our proposed explanations methodology builds on the framework for Shapley values, and Shapley values for feature importance, we provide a detailed theoretical description of the Shapley value concept in the following.

## 2.2 Shapley values of a coalitional game

A coalitional game $\langle N, v\rangle$ consists of a finite set of players $N$, and a value function $v : 2^{|N|} \rightarrow \mathbb{R}$ which maps a coalition $S \subseteq N$ of players to the real numbers, such that $v(\emptyset) = 0$. $N$ denotes the *grand* coalition of all players. We also assume that the players not belonging to a coalition $S$ do not have any influence on $v(S)$. The value function $v(S)$ describes how much collective payout a set of players can gain by forming the coalition $S$.

A *solution* of a game $\langle N, v\rangle$ is a mapping that assigns to each player her expected marginal contribution, that is splitting the worth of $v(N)$ among the players in a "fair" way. In general, the marginal contribution of a player depends on the order in which she joins the coalition (Çetiner 2013). Depending on how we define "fair", different solution concepts are preferred. Çetiner (2013) provides good explanations to most common concepts, including *the Core* and variants of this, *the Nucleolus*, *the Kernel*, the *Owen set* and *the Shapley value*. In this paper, we devote our attention to the latter solution. The Shapley value was introduced by Shapley (1953), and it has a set of desirable properties as we will see below.

Shapley (1953) expresses the Shapley value of player $i$ in a coalitional game $\langle N, v\rangle$ as

$$\varphi_i = \sum_{S\subseteq N\setminus\{i\}} \frac{|S|!\big(|N|-|S|-1\big)!}{|N|!}\cdot\big[v(S\cup\{i\})-v(S)\big] \tag{1}$$

where $|N|$ is the total number of players, $|S|$ denotes the number of players in coalition $S$, and $v(S)$ describes the total expected sum of payouts the members of $S$ can obtain by cooperation. The sum extends over all subsets $S$ of $N\setminus\{i\}$. We also define the non-distributed gain $\varphi_0 = v(\emptyset)$, which describes the fixed payoff which is not associated to the actions of any of the players, although this is often zero for coalitional games (Aas et al 2021).

The Shapley value of a player is the average of its marginal contributions with respect to all the permutations. Hence, an alternative expression of the Shapley value of player $i$ in a coalitional game $\langle N, v\rangle$ is

$$\varphi_i = \frac{1}{|N|!}\sum_{\mathcal{O}\in\pi(|N|)} \big[v\big(\mathrm{Pre}^i(\mathcal{O})\cup\{i\}\big)-v\big(\mathrm{Pre}^i(\mathcal{O})\big)\big], \tag{2}$$

where $\pi(|N|)$ is the set of all permutations of $|N|$ elements, and $\mathrm{Pre}^i(\mathcal{O})$ is the set of all players which precede the $i$-th player in permutation $\mathcal{O}\in\pi(|N|)$. For more details, see Çetiner (2013), Castro et al (2009) and Štrumbelj and Kononenko (2011).

Shapley (1953) shows that the Shapley value is the unique solution which satisfies the following properties:

*Efficiency:* The total gain is distributed:

$$\sum_{i=0}^{|N|}\varphi_i = v(N) \tag{3}$$

*Symmetry:* If $i$ and $j$ are two actors who are equivalent in the sense that

$$v(S\cup\{i\}) = v(S\cup\{j\}) \tag{4}$$

for every subset $S$ of $N$ which contains neither $i$ nor $j$, then $\varphi_i = \varphi_j$.

*Linearity:* If two coalition games described by value functions $v$ and $w$ are combined, then the distributed gains should correspond to the gains derived from $v$ and the gains derived from $w$:

$$\varphi_i(v+w) = \varphi_i(v)+\varphi_i(w) \tag{5}$$

for every $i\in N$. Also, for any real number $a$

$$\varphi_i(av) = a\varphi_i(v) \tag{6}$$

for every $i \in N$.

*Zero player (null player):* $\varphi_i = \varphi_0$ iff player $i$ is a null-player, i.e. $v(\{i\}) = \varphi_0$ and $v(S \cup \{i\}) = v(S)$ for all coalitions $S \in N$. Here, $\varphi_0 = v(\emptyset)$ is define as the non-distributed gain which describes the fixed payoff which is not associated to the actions of any of the players. For coalitional games this is often zero (Aas et al 2021).

### 2.3 Shapley values for feature importance

Lipovetsky and Conklin (2001) apply Shapley values to determine the comparative usefulness of features/regressors in multiple regression analysis, specifically focusing on the difficulties due to multicollinearity among features. Shapley values are also applied by Štrumbelj and Kononenko (2010) to quantify the comparative importance of features, with focus on explaining individual predictions produced by classification models. They propose a *sampling-based* method to approximate the Shapley values to overcome the initial exponential time complexity. Štrumbelj and Kononenko (2011) adapt the explanation method for use with regression models. Lundberg and Lee (2017) propose an alternative approximation method called the *Kernel SHAP*. According to the authors, this method can improve the sample efficiency of the model-agnostic estimators by restricting attention to specific model types, and develop faster model-specific approximation methods. Aas et al (2021) extend the *Kernel SHAP* method to handle dependent features.

In the following we briefly review the *sampling-based* explanation method proposed by Štrumbelj and Kononenko (2011), to efficiently calculate the Shapley value for feature importance in a regression model. See Lipovetsky and Conklin (2001) and Štrumbelj and Kononenko (2010, 2011) for details.

We consider a standard machine learning setting where a training set $\mathcal{D}^{train}$, consisting of $J$-dimensional feature vectors and corresponding observed responses, is used to train a predictive model $f$. Let the feature space be defined as $\mathcal{A} \in \mathcal{A}_1 \times \mathcal{A}_2 \times \cdots \times \mathcal{A}_J$, and let $p$ be the probability mass function defined on $\mathcal{A}$. Here, we assume that individual features are mutually independent. For the dependent case, see Aas et al (2021). Now let the features in such a model act as players in the game defined in Sect. 2.2. The aim is to express how each feature affects the prediction of a model $f : \mathcal{A} \to \mathbb{R}$ in a specific test data point $x \in \mathcal{A}$. Let the contribution of a subset of feature values in this specific data point be the expectation caused by observing those feature values. Formally, the value function is given as

$$v(S)(x) = \sum_{z \in \mathcal{A}} p(z)\big(f(\tau(x, z, S)) - f(z)\big), \tag{7}$$

where $\tau(x, z, S) = (u_1, \ldots, u_J)$ such that $u_j = x_j$ iff $j \in S$ and $u_j = z_j$ otherwise. The $x$ values are the true explanatory variables of the investigated data point, while $z$ are random data points from the feature space $\mathcal{A}$. For simplicity, assume that $\mathcal{A}$ is discrete. In the continuous case, the second sum in the following expression is replaced by an integral. The Shapley value Eq. (2)] for the $j$-th feature of the game $\langle N, v\rangle$, with

$v$ defined in Eq. (7), is now

$$\varphi_j(x) = \frac{1}{J!} \sum_{\mathcal{O} \in \pi(J)} \sum_{z \in \mathcal{A}} p(z) \Big[ f(\tau(x, z, \text{Pre}^j(\mathcal{O}) \cup \{j\})) - f(\tau(x, z, \text{Pre}^j(\mathcal{O}))) \Big], \tag{8}$$

where $\pi(J)$ is the set of all permutations of the $J$ different features, and $\text{Pre}^j(\mathcal{O})$ is the set of all features which precede the $j$-th feature in permutation $\mathcal{O} \in \pi(J)$. Note that the term $f(z)$ occurs for both $v(\text{Pre}^j(\mathcal{O} \cup \{j\})$ and $v(\text{Pre}^j(\mathcal{O}))$, hence they cancel.

To calculate an exact Shapley value, all possible coalitions have to be evaluated with and without the $j$-th feature (Molnar 2021, Ch. 5). Since we do not know the distribution $p(z)$, computing $v(S)$ is difficult. Furthermore, the number of possible coalitions of a set $N$ of $|N|$ features is $2^{|N|}$. Hence, finding the exact solution becomes impossible, except with very few features. However, the Shapley values in the form presented in Eq. (8) facilitate the use of random sampling and an efficient approximation algorithm. See Castro et al (2009) and Štrumbelj and Kononenko (2010, 2011) for details. The approximated Shapley value for feature importance is given as

$$\hat{\varphi}_j(x) = \frac{1}{M} \sum_{m=1}^{M} \Big[ f(\tau(x, z^m, \text{Pre}^j(\mathcal{O}^m \cup \{j\}))) - f(\tau(x, z^m, \text{Pre}^j(\mathcal{O}^m))) \Big], \tag{9}$$

where for each sample $m$, a permutation $\mathcal{O} \in \pi(|N|)$ and a point $z^m \in \mathcal{A}$ are sampled according to $p$. Since $p$ is usually unknown, in practice this means resampling from a dataset, as described by Štrumbelj and Kononenko (2010, 2011). In this way, $\hat{\varphi}_j(x)$ approximates how the prediction of the data point of interest, $x$, depends on the $j$-th feature.

## 3 Shapley values for cluster importance

To understand and interpret how a model produces a prediction for a specific data point, the above Shapley value for feature importance is a useful measure. In addition to such feature importance, it is essential to understand the data used to train the model, and to understand how the data affects the model's predictions. We propose to obtain a measure of the importance of various clusters of the training data, by letting the different clusters of the data take part as players in a game where the predictions are the payouts.

With clusters we intend some kind of sub division of the training data, with the extreme case having each individual observation in separate clusters. Given the problem at hand, there will most often exist natural divisions of the training data that leads to meaningful clusters and hence explanations. These could for example be time periods for the data collection, stratification based on covariates that are not part of the black-box model because they are not legal to use, or simply not used, etc. Of course, a random sub division of the training data is also possible, but the interpretation of the results becomes less interesting. We give some examples of meaningful clusters in the next section.

As in the previous section, we consider a regression function $f : \mathcal{A} \to \mathbb{R}$, where $\mathcal{A} \in \mathcal{A}_1 \times \cdots \times \mathcal{A}_J$. Now, we divide the training dataset into $K$ disjoint clusters $Q_k$, such that $Q_1 \cup \cdots \cup Q_K$ is equal to the full training dataset $\mathcal{D}^{train}$. We let the different clusters $Q_k$ be the players in the game $\langle N, v\rangle$. As before, we let $N$ be the grand coalition, which means that $N$ is the dataset which contains all clusters, and hence $N = \mathcal{D}^{train}$. We let $S \subseteq N$ denote coalitions of clusters of the training data.

The aim is to investigate how the learning process of the model is affected by the different clusters of the training data. That is, for a new data point $x \in \mathcal{A}$, we are interested in how the data in cluster $Q_k$ contributes to the prediction of $f(x)$. Hence, we define the game $\langle N, v\rangle$ with value function

$$v(S)(x) = f_S(x), \tag{10}$$

where $f_S$ is a function which is trained on a dataset composed by the union of $Q_k$ for $k \in S$. We suggest to let the Shapley value for the $k$-th cluster of the game $\langle N, v\rangle$ with value function defined in Eq. (10) expressed on the form Eq. (2) be

$$\varphi_k(x) = \frac{1}{K!} \sum_{\mathcal{O} \in \pi(K)} \Big( f_{\text{Pre}^k(\mathcal{O} \cup \{k\})}(x) - f_{\text{Pre}^k(\mathcal{O})}(x) \Big), \tag{11}$$

where $\pi(K)$ is the set of all permutations of $K$ clusters, and $\text{Pre}^k(\mathcal{O})$ is the set of all clusters which precede the $k$-th cluster in permutation $\mathcal{O} \in \pi(K)$.

When we have no data, that is when $S = \emptyset$, we usually define the predictions to be 0, that is $f_\emptyset(x) = 0$ for all $x \in \mathcal{A}$. This also ensures that $v(\emptyset) = 0$. We interpret the Shapley value of the $k$-th cluster, $\varphi_k$, as how much the $k$-th cluster contributes to increase or decrease the prediction relative to 0. In most cases, we find this interpretation most intuitive. However, in cases where we have prior knowledge about the distribution of the response $y$, it might be beneficial to set $f_\emptyset$ equal to the mean, say, of that distribution. Alternatively, we can pre-process the training data, and center it at 0.

Following the same arguments as for the approximation of the Shapley value for feature importance, a sampling based approximation of the Shapley value for cluster importance is

$$\hat{\varphi}_k(x) = \frac{1}{M} \sum_{m=1}^{M} \Big( f_{\text{Pre}^k(\mathcal{O}^m \cup \{k\})}(x) - f_{\text{Pre}^k(\mathcal{O}^m)}(x) \Big), \tag{12}$$

where for each sample $m$, a permutation $\mathcal{O}^m \in \pi(K)$ is randomly drawn (uniformly).

Other approximations than Eq. (12) could be suggested, and the statistical properties should be studied and compared. We proceed with the above approximation in this paper, and show empirically that an approximation of the form of Eq. (12) works excellently on a set of small examples where it is possible to compute exact Shapley values. The implementation is described in Algorithm 1.

**Algorithm 1:** Approximated Shapley value for the importance of cluster $k$ for data point $x$.

**Required:**
Number of iterations $M$;
**Initialization:**
Divide training data into clusters: $Q_1, Q_2, \ldots, Q_k$;
$\varphi_k(x) := 0$;
**for** $m = 1, \ldots, M$ **do**
    Sample a random permutation $\mathcal{O} \in \pi(K)$;
    Form dataset $\mathcal{D}^+$ consisting of $Q_k$ and $Q_i$ for $i$ which precede $k$ in $\mathcal{O}$;
    Use dataset $\mathcal{D}^+$ to train a function $f_{\mathcal{D}^+}$;
    Form dataset $\mathcal{D}^-$ consisting of $Q_i$ for $i$ which precede $k$ in $\mathcal{O}$;
    Use dataset $\mathcal{D}^-$ to train a function $f_{\mathcal{D}^-}$;
    Update Shapley values: $\varphi_k(x) := \varphi_k(x) + f_{\mathcal{D}^+}(x) - f_{\mathcal{D}^-}(x)$
**end**
$\varphi_k(x) := \frac{\varphi_k(x)}{M}$;

### 3.1 Computational effort

When approximating the Shapley values using the sampling procedure in Algorithm 1, the model is retrained for each sample $m \in \{1, \ldots, M\}$. The effort is, however, usually significantly smaller than training the original model, because the size of the various datasets, which depends on the size of the coalition $S^m \subseteq N$, is significantly reduced for many of the samples. Nevertheless, the proposed method is computationally expensive. Fortunately, the retraining can be done in parallel. It is also possible to utilize the property that the models are trained on unions of clusters that are order independent. Furthermore, the retraining process does not need to be performed repeatedly for each new test point $x \in \mathcal{D}^{test}$. When a model is trained, it can be reused when explaining a new prediction. Furthermore, in our experience, the approximation of Shapley values rapidly converge.

## 4 Examples and demonstrations

First in this section, we discuss a simple price estimation problem, and illustrate how we can use Shapley value for cluster importance to better understand the predictions of the model. To be able to verify and understand the Shapley values, the first example is deliberately very simple, and exact values can be calculated in some cases. Secondly, we present an example where we use Shapley values for cluster importance to reveal that a predictor is biased. The third example illustrates, on a real, publicly available dataset, how Shapley values for cluster importance can supplement feature importance measures, providing insight not only into the importance of a feature, but also how this feature affects predictions. The example also illustrates that the explanations produced using Shapley values for cluster importance correspond to our intuitive explanations for easily interpretable prediction models.

### 4.1 Illustrative example with exact solution

The following three prediction models all use previous sales to predict the sales price of a car:

$f$: average sales price of all previous sales,
$g$: average sales price of similar cars,
$h$: black-box model trained on previous sales data.

One would typically argue that $f$ is transparent, what similar cars mean is not revealed in $g$, and the inner workings of $h$ is hidden from us. Although we know the inner workings of $f$, we need access to previous sales (the model's training data) before we can say anything about its predictions. It would for example be relevant to disclose how different data-points contribute to the prediction. If the prediction is based on average sales price as in $f$, each data-point contribute $1/n$, but what is the contribution of a data-point when the prediction is based on a black-box model as in $h$? And which data-points are most influential? When the dataset is large, it is impractical to treat every data-point individually. Therefore, we cluster the training data into meaningful clusters, and quantify how different clusters affect the prediction. It can for example be interesting to cluster the training data based on car type and calculate the associated Shapley values for cluster importance. Clustering the data based on weekday of sale, is probably less interesting (unless for some reason people tend to pay more on certain weekdays). Note that this type of information is relevant for all of the three prediction models, independent of whether or not we understand the inner workings of the model.

In the following we describe a simple regression problem on a dataset comprising one explanatory variable and one response $y$, as illustrated in Fig. 1a. The example is generic, but we can think of the response as sales price of a car, and the explanatory variable as engine power. We show how Shapley values for cluster importance can contribute in interpreting the regression models and explaining its predictions.

#### 4.1.1 Linear regression

First, we train a linear regression model $f = ax + b$ using 18 datapoints (The dataset is provided in "Appendix C"). The fitted model and the data is illustrated in Fig. 1, and the estimated parameters are $\hat{a} = 1.22$, and $\hat{b} = 2.84$. Given a new observation $x = 4$, this model produces a prediction $\hat{y} = \hat{a}x + \hat{b} = 7.72$, shown in black.

Suppose now that the training data is divided into two clusters; cluster 1 (blue) and 2 (green) as shown in Fig. 1b (Cluster 1 and 2 can for example comprise cars from Italy and Germany respectively). We can now quantify how different clusters affect our prediction $\hat{y}$ using Shapley values for cluster importance. We calculate the exact Shapley values by averaging over the marginal distributions with respect to all the permutations (see Eq. 2). To do this we need to train two new models using data from cluster 1 and 2; $f_1 = a_1x + b_1$ and $f_2 = a_2x + b_2$. In this example $\hat{a}_1 = 1.20$, $\hat{b}_1 = 7.14$, $\hat{a}_2 = 1.19$, and $\hat{b}_2 = 0.83$. We define the non-distributed gain, $\varphi_0$ to be 0, meaning that when we have no data we let the predictions be 0 ($f_\emptyset = 0$). We have $K = 2$ clusters which gives $2! = 2$ possible permutations. These are listed in the first column of Tables 1, 2. In the second and third column of the tables, the accompanying clusters $S \cup \{k\}$ and $S$ are listed. In column four and five, the exact predictions are expressed, and difference between them are expressed in the sixth column. Finally, we calculate the Shapley value for the importance of cluster 1 and 2 by averaging over the values in column six of Table 1 and 2 respectively: $\varphi_1(x) = 7.03$ and $\varphi_2(x) = 0.68$. Note that the sum of the Shapley values equals the predicted value ($y = 7.72$), and hence the efficiency property holds, (see Eq. 3), that is that the total gain is distributed, $\sum_{i=0}^{|N|} \varphi_i = v(N)$.

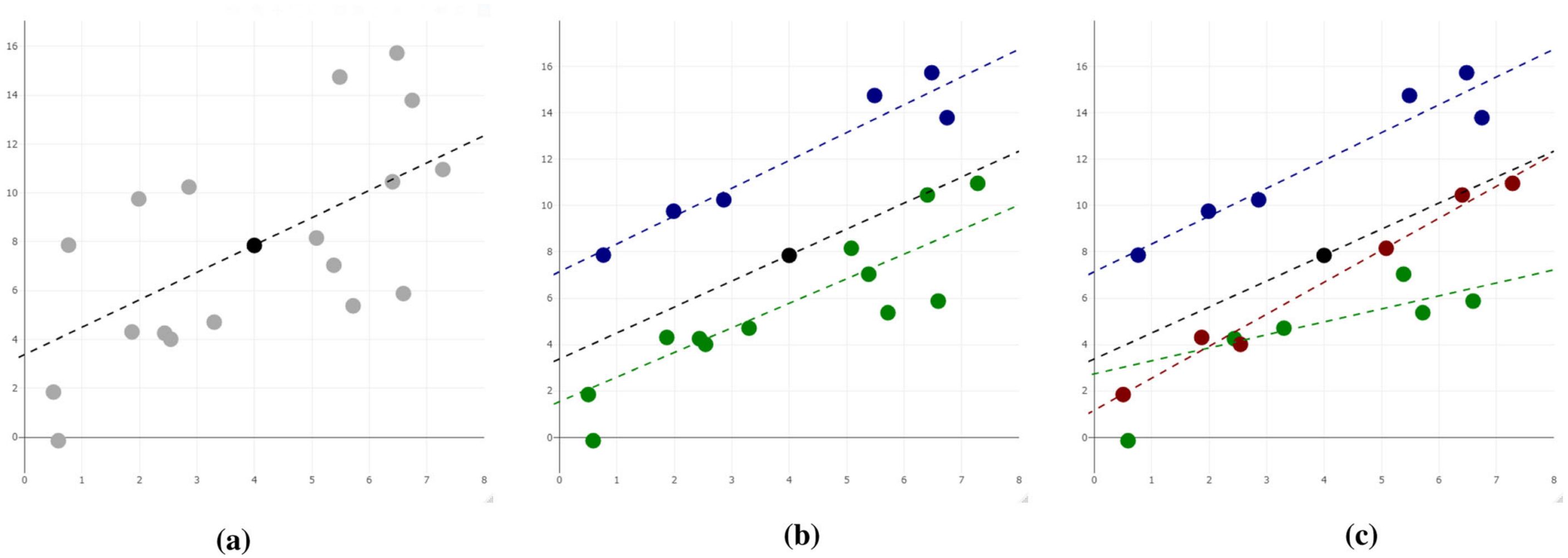

**Fig. 1** **a** Displays a set of datapoints and a linear model $f = ax + b$ trained on the full dataset. A prediction of $f(x)$ for $x = 4$ is shown in black. In **b**, the training dataset is divided into two clusters: cluster 1 (blue) and cluster 2 (green). In **c**, the training dataset is divided into three clusters: cluster 1 (blue), cluster 2 (red) and cluster 3 (green) (Color figure online)

**Table 1** *Cluster 1:* Calculation of the exact Shapley value for cluster importance of cluster 1: $\varphi_1 = 1/2(11.95 + 2.12) = 7.03$

| $\mathcal{O}$ | $S \cup \{k\}$ | $S$ | $f_{\mathcal{O} \cup \{k\}}$ | $f_{\mathcal{O}}$ | $f_{\mathcal{O} \cup \{k\}} - f_{\mathcal{O}}$ |
|---|---|---|---|---|---|
| {1 2} | {1} | Ø | 11.95 | 0 | 11.95 |
| {2 1} | {1, 2} | {2} | 7.72 | 5.60 | 2.12 |

**Table 2** *Cluster 2:* Calculation of the exact Shapley value for cluster importance of cluster 2: $\varphi_2 = 1/2(-4.23 + 5.60) = 0.68$

| $\mathcal{O}$ | $S \cup \{k\}$ | $S$ | $f_{\mathcal{O} \cup \{k\}}$ | $f_{\mathcal{O}}$ | $f_{\mathcal{O} \cup \{k\}} - f_{\mathcal{O}}$ |
|---|---|---|---|---|---|
| {1 2} | {1, 2} | {1} | 7.72 | 11.95 | −4.23 |
| {2 1} | {2} | Ø | 5.60 | 0 | 5.60 |

When interpreting the explanations, the analogy to coalitional game theory is useful. The Shapley value of a coalitional game fairly distributes the payouts of a game between the cooperating players. Here, clusters are the players and predictions are the payouts. Hence, a Shapley value quantifies the contribution of a cluster. Cluster 1's contribution is 7.03 while cluster 2's contribution is 0.68, assuming that the prediction is 0 when we have no training data.

Suppose now that the second cluster can be further divided, giving us a training data set with three meaningful clusters (cluster 1, 2 and 3) as illustrated in Fig. 1c (for example different car brands; Ferrari, Audi, Volkswagen). Again, we can calculate the exact Shapley values by averaging over the marginal distributions with respect to all permutations. With three clusters, we have $3! = 6$ possible permutations. These are listed in the first column of Tables 4, 5, 6 in "Appendix A". The clusters $S \cup \{k\}$ and $S$ are listed in the second and third column and in column four and five, the exact predictions are written. The difference between them are expressed in the sixth column, and the average of these gives the Shapley value for cluster importance: $\varphi_1(x) = 5.95$, $\varphi_2(x) = 1.54$ and $\varphi_3(x) = 0.23$.

We can easily check that the efficiency property (see Eq. 3) holds, that is that the total gain is distributed. This means that the sum of the Shapley values ($5.95 + 1.54 + 0.23 = 7.72$) equals the prediction ($f(4) = a \cdot 4 + b = 1.22 \cdot 4 + 2.84 = 7.72$).

As long as the number of clusters $k$ is small, it is possible to calculate the exact Shapley values. However, as the number of permutations is $k!$, this becomes intractable for large $k$. Hence, the approximation is essential. We illustrate this for the linear model, and compare the results. The values are displayed in Fig. 2 as the number of iterations grows from 1 to 250. $\varphi_1$ are shown in blue, $\varphi_2$ in red and $\varphi_3$ in green. The dotted lines shows the exact values as calculated above.

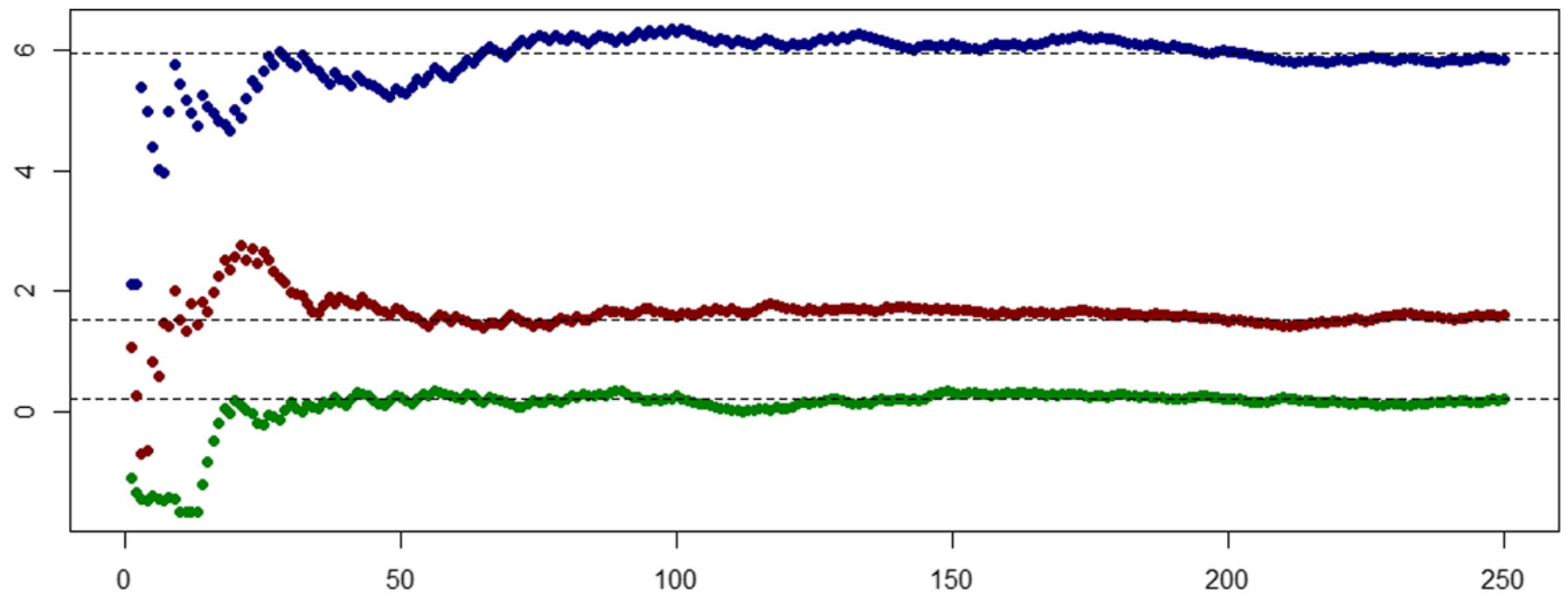

**Fig. 2** Approximate Shapley values for the *linear regression* model when number of iterations grows from 1 to 250 ($\varphi_1$ in blue, $\varphi_2$ in red and $\varphi_3$ in green). The dotted lines shows the exact values (Color figure online)

#### 4.1.2 *n* clusters

For illustrative purposes, we can let each data point have its own cluster, that is $k = n$ clusters with 1 element in each cluster. Approximated Shapley values for each cluster (comprising one data point) are shown in Fig. 3a. As expected, we observe that the data points with large response values contribute the most to increase predictions (relative to 0), while data points with low response contribute less, and some data points have a negative contribution.

For comparison, we display Cook's distances for each data point in Fig. 3b. Cook's distances are commonly used for judging the influence of data points in the parameter vector estimation in least squares regression (Cook 1977, 1979; Kumar et al 2019; Kannan and Manoj 2015). Using Cook's distances, the influence of the $i$-th data point is given as

$$D_i = \frac{\sum_{j=1}^{n}(f_N(x_j) - f_{N\setminus\{i\}}(x_j))^2}{p \cdot MSE}, \tag{13}$$

where observation $i$ is excluded when fitting $f_{N\setminus\{i\}}$, $p$ is the number of coefficients in the regression model, and $MSE$ is the mean squared error.

#### 4.1.3 Nearest neighbor

We now return to the case with three clusters (Fig. 1c), and replace the linear model with a $k$-nearest neighbor model. We use the Fast Nearest Neighbor Search Algorithms and Applications {FNN} (Beygelzimer et al 2019) implementation in R (R Core Team 2019), with $k = 1$ neighbor, and the *kd_tree* nearest neighbor search algorithm. For the new data point $x = 4$, this model selects the nearest point in the training data, and outputs the response value of this datapoint. For example, the nearest point in cluster 1 is $(x, y) = (2.86, 10.24)$ Hence, the prediction $\hat{y} = 10.24$. The nearest point in $S_{1,3}$ (the subset which comprise points from cluster 1 and from cluster 3) is $(x, y) = (3.30, 4.71)$, and hence the prediction is $\hat{y} = 4.71$. The exact calculations of all permutations of clusters 1, 2 and 3 are shown in Table 7, 8, 9. Approximated values are shown in Fig. 4 as number of iterations grow from 1 to 250, together with the exact values shown as horizontal lines, indicating that the approximated values quickly converge to the correct values.

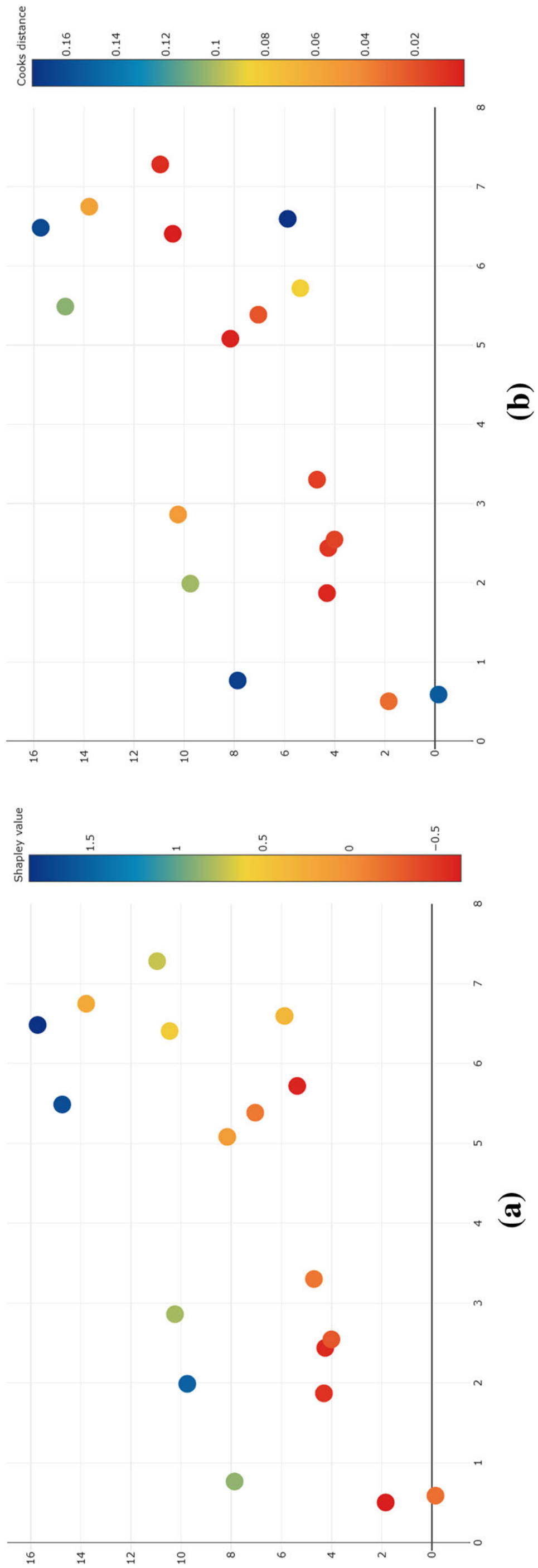

**Fig. 3** **a** Displays the Shapley value for each cluster (comprising only one data point each). In **b**, Cook's distances are displayed for each data point

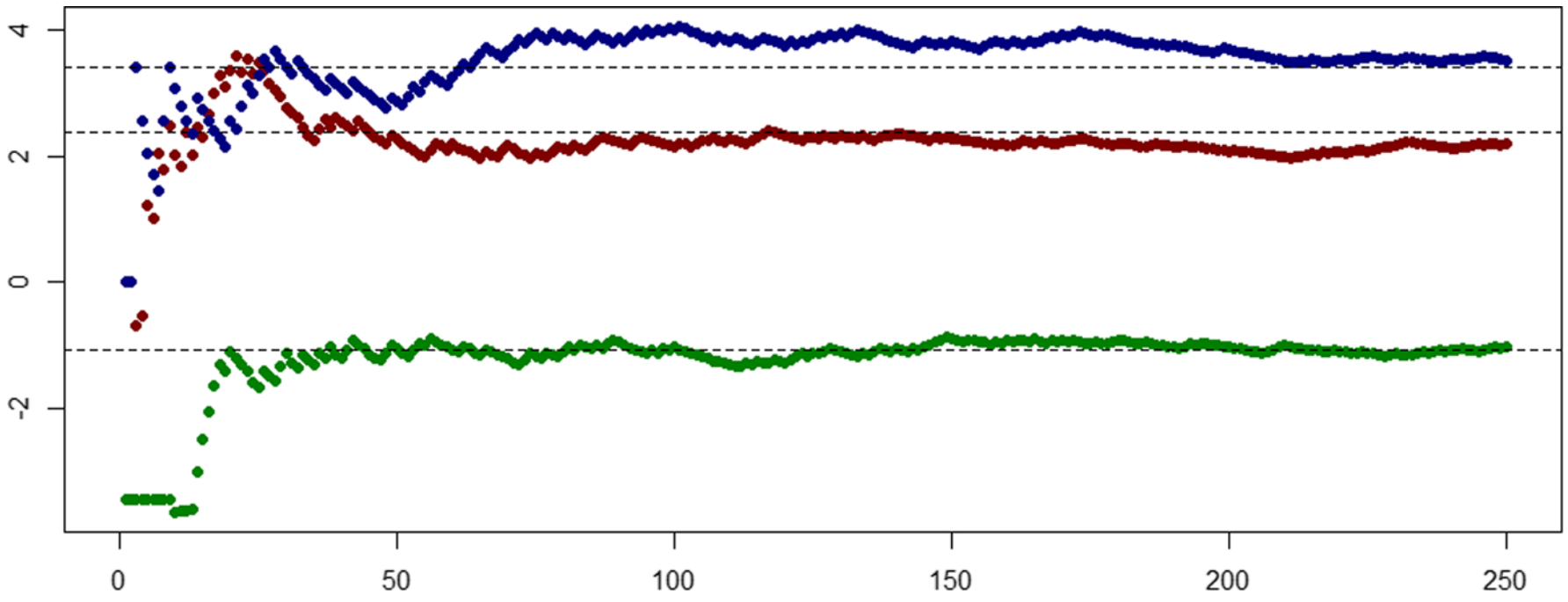

**Fig. 4** Approximate Shapley values for the *k*NN regression model when number of iterations grows from 1 to 250 ($\varphi_1$ in blue, $\varphi_2$ in red and $\varphi_3$ in green). The dotted lines shows the exact values

#### 4.1.4 Black-box models

The simple models we have explored above, linear models and nearest neighbor models, are (to some extent) interpretable and it is fairly easy to predict and explain their predictions. Our focus has been concentrated on these simple models in order to validate that the information conveyed through Shapley values correspond to our intuition and understanding. However, the use of Shapley values becomes relevant in cases where we cannot interpret the models directly, which is the case in black box models.

Figure 5a and b show the Shapley values when applied to a random forest model with 10 trees and maximum nodes set to 5 (Liaw and Wiener 2002) and a support vector machine model with default setup (Meyer et al 2021) respectively. The Shapley values for training data importance gives us information about how the instances of the different clusters of the training dataset contribute to the prediction, even when the predictor is a black box. In our example, the training data from cluster 1 (shown in blue) comprises Ferrari sales data. As expected, this training data cluster contributes to increase the prediction. This is true for both the random forest model and the support vector machine, with a slightly higher importance in the second model. The importance of the second cluster (shown in red) is approximately equal in the two models. Note that a good prediction model would of course include car-brand as an explanatory variable if this information is available. However, not all models are good, and information about the importance of different clusters can allow the users to question the reasoning of the model. Furthermore, the user may possess information that is not available to the model, and sometimes the model should not be allowed to use all types of information. The latter is the topic of the following example.

### 4.2 Revealing biased behaviour

Recent studies demonstrate that machine learning algorithms can reproduce and amplify biases from the real world (Buolamwini and Gebru 2018). For example,

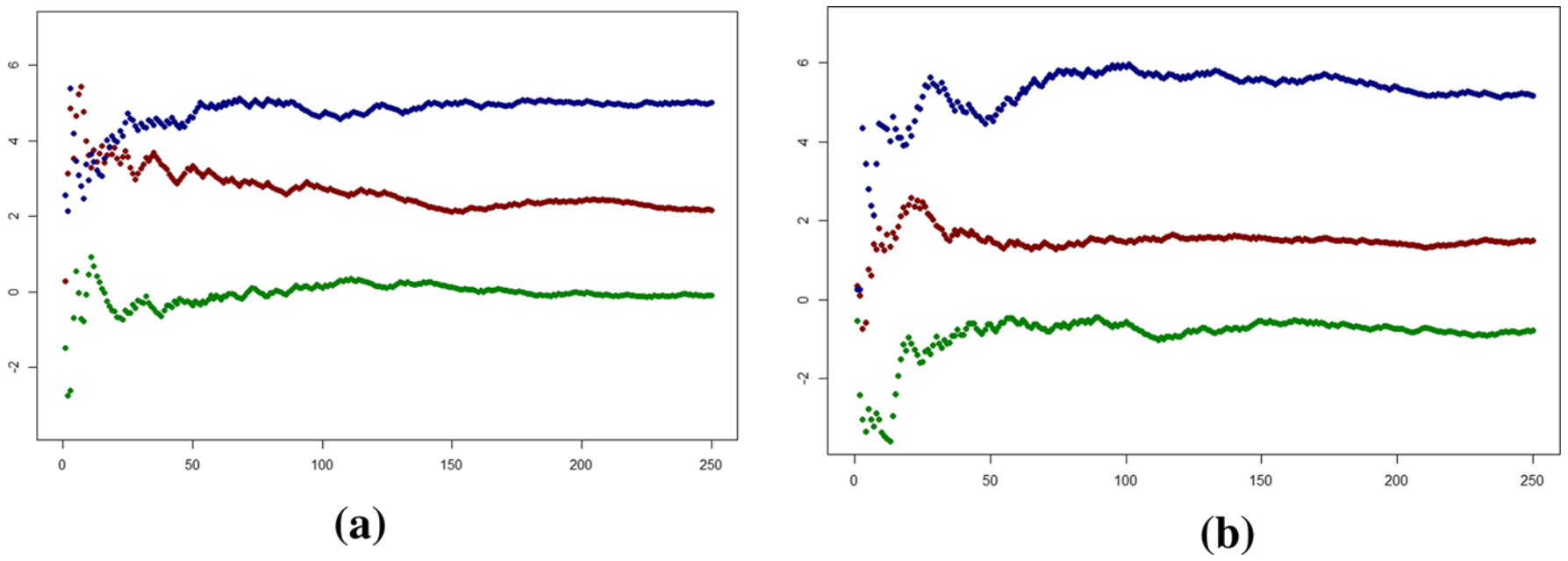

**Fig. 5** Approximate Shapley values for the *random forest* **a** and *support vector machine* **b** regression model when number of iterations grows from 1 to 250 ($\varphi_1$ in blue, $\varphi_2$ in red and $\varphi_3$ in green) (Color figure online)

Angwin et al (2016) report that a software used across the United States to predict future criminals has racial bias. Similarly, Lum and Isaac (2016) demonstrate that predictive policing of drug crimes, used by law enforcement to try to prevent crime before it occurs, results in increasingly disproportionate policing of historically over-policed communities.

In this section we consider how explanations based on Shapley values for cluster importance can be used to analyse and investigate if a model is discriminative. We consider a simulated example where an algorithm determines the size of a loan a customer is granted by a bank. Suppose the customer wants to know if and how her country of birth affects the decision. Obviously, if a model uses country of birth as a feature, it is easy to calculate and use the Shapley values for feature importance to explain how this affects the predictions. However, to avoid making the algorithm discriminative, country of birth is typically excluded as a feature. Nevertheless, a prediction can rely on national origin indirectly through other hidden dependencies, such as for example residential area.

Let the size of the granted loan be given by $f : \mathcal{A} \to \mathbb{R}$, where the feature space $\mathcal{A} \in \mathcal{A}_1 \times \cdots \times \mathcal{A}_J$. In addition to the explanatory variables, $x_i$, for $i = 1, \ldots, J$, we define a categorical variable, $x_D$, which denotes a discriminative property; in this example country of birth. In the numerical results presented below, we simulate a training and a test dataset comprising 100 instances from each country, such that both the training and test dataset comprise 300 instances in total. Furthermore, we use four explanatory variables ($J = 4$). Based on $x_D$, we cluster the dataset into 3 different clusters (country $A$, $B$ and $C$), and use Shapley values to quantify the importance of each cluster.

### 4.2.1 Response and explanatory variables independent on the discriminative property

As a baseline, we first define the process generating the response to be white noise, that is

$$y = \epsilon \qquad \text{where } \epsilon \sim N(0, 1). \tag{14}$$

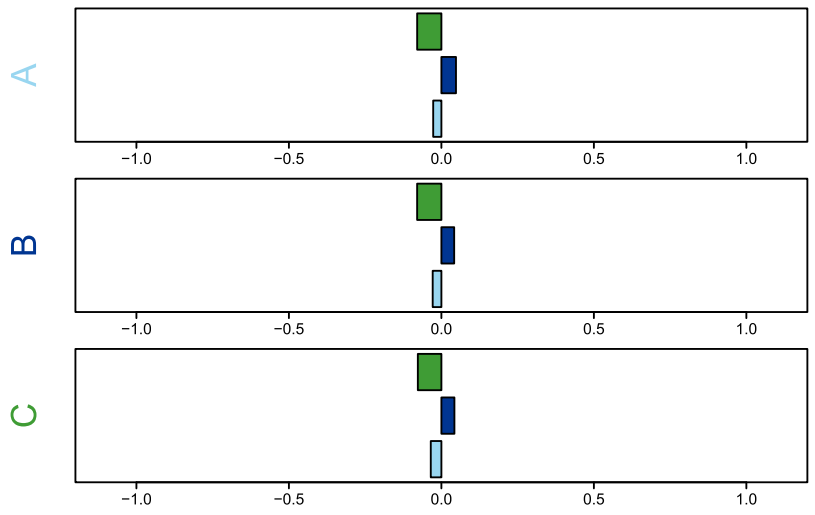

**(a)** $y$ independent on $x_D$, $x$ independent on $x_D$

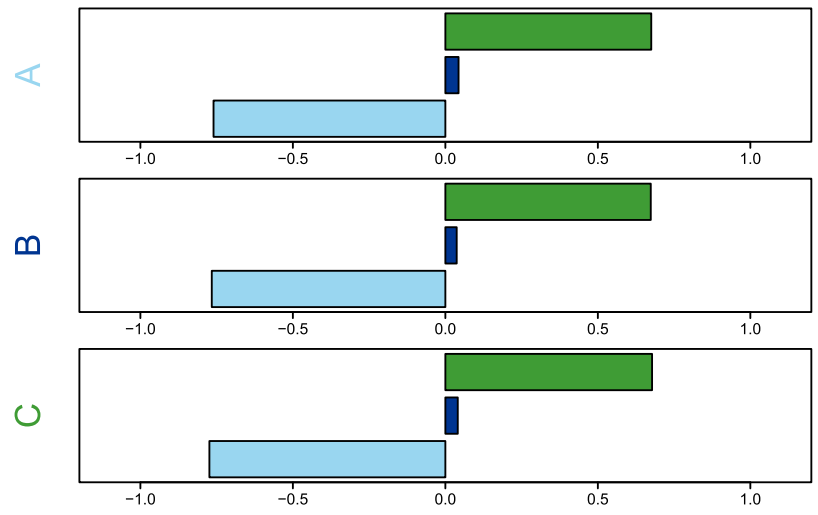

**(b)** $y$ dependent on $x_D$, $x$ independent on $x_D$

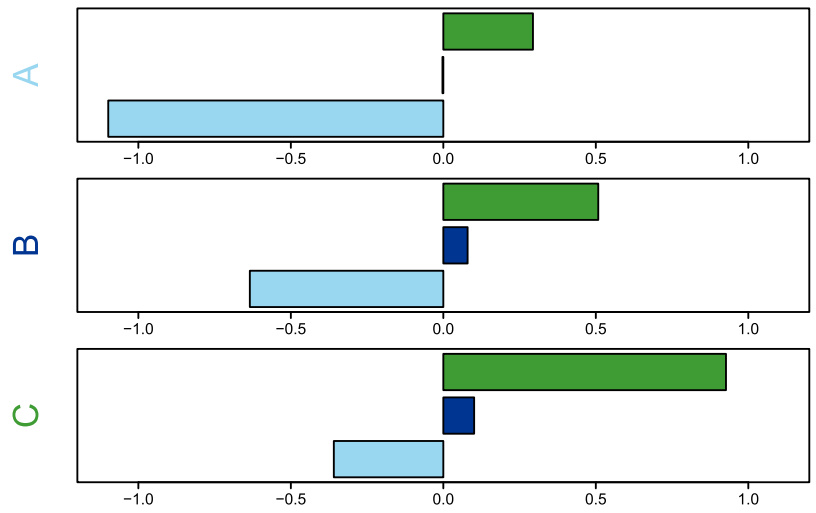

**(c)** $y$ dependent on $x_D$, $x$ dependent on $x_D$

**Fig. 6** Shapley values for all the test points in a country are calculated, and the average Shapley values for that country is presented. **a**–**c** shows results from calculations described in Sects. 4.2.1, 4.2.2 and 4.2.3 respectively. Results for individuals of the *test* data from country $A$, $B$ and $C$ are shown in the upper, middle and lower subplots respectively. The Shapley values for training data cluster importance of the three countries $A$, $B$ and $C$ are shown in light blue, blue and green respectively

Even if the explanatory variables are not involved, we generate $x_1, \ldots, x_4$ also as iid $N(0, 1)$ variables, and use the training dataset with these covariates and this response to train a $k$ nearest neighbor model with $k = 10$.

No matter which model we use, if it is trained on this dataset, it will of course not discriminate based on $x_D$ (country of birth), because both the explanatory variables, $x_i$, and the response, $y$, are independent on $x_D$ and on each others. Hence, if we explain the predictions for a set of individuals, we expect the average Shapley values for training data cluster importance to be approximately zero. We observe this in the three barplots in Fig. 6a. Here, the Shapley values for all predictions for the 100 instances belonging to a country are calculated, and the average Shapley values for individuals in the *test* dataset belonging to country $A$, $B$ and $C$ are shown in the upper, middle and lower subplot respectively. The Shapley values for cluster importance are shown in light blue, blue and green. These values describe the importance of the three different clusters of the *training* data, comprising individuals from country $A$, $B$ and $C$ respectively.

### 4.2.2 Response is dependent on the discriminative property, but explanatory variables are independent

We now change the response in the training dataset such that the response deterministically depends on the sensitive information $x_D$ (country of birth), by letting

$$y = x_D + \epsilon, \tag{15}$$

where the explanatory variables $x_1, \ldots, x_4$ are as defined above, and $y$ is independent of these. We let $x_D$ take values $-1$, 0 and 1 for country $A$, $B$ and $C$ respectively.

As in Sect. 4.2.1, we use a $k$NN model with $k = 10$, now trained on a dataset with the new response values generated by (15). The Shapley values for training data cluster importance for the predictions using the new responses are displayed in Fig. 6b. The light blue bars show that individuals from country $A$ contribute to decrease the predictions, while individuals from country $C$ contribute to increase predictions. But this does not indicate that the model is discriminative. The explanatory variables $x_1, \ldots, x_J$ are drawn from a standard normal distribution, and hence, all the explanatory variables are independent of country of birth ($x_D$), and therefore the model cannot take country of birth into account. We observe that the three plots are almost identical, indicating that the individuals in the different groups ($A$, $B$ and $C$) are treated equally by the model.

### 4.2.3 Response and explanatory variables dependent on the discriminative property

However, if we include dependence between $x_D$ and the explanatory variables, the model might be discriminative. In the following, we once again use the response values generated by (15). But now, we alter the explanatory variables $x_1, \ldots, x_J$ such that they are dependent on $x_D$, in the following way

$$\begin{aligned} x_1 &\sim N(x_D, 1) \\ x_2 &\sim N(-x_D, 1) \\ x_3 &\sim N(2x_D, 1) \\ x_4 &\sim N(-2x_D, 1). \end{aligned} \tag{16}$$

In the bank loan setting, this mimics that country of birth affects some of the covariates, as well as the size of the loan given in the training data. The results are displayed in Fig. 6c. We observe that predictions for individuals from country $A$ ($x_D = -1$) are severely reduced by individuals from this country (light blue). Individuals from this country also contribute to reduce the predictions of individuals from the other countries, but the reduction is smaller. Similarly, individuals from country $C$ ($x_D = 1$) contribute to increase the predictions of individuals from country $C$ more than individuals from the two other countries. Unlike in Fig. 6b, the subsets of the training data now affect individuals from the three countries differently, and this practice can be regarded as discriminative.

**Table 3** Features used to predict the daily count of rented bikes

| Name | Description |
|---|---|
| Season | 1: winter, 2: spring, 3: summer, 4: fall |
| Weekday | Day of the week |
| Weathersit | 1: clear/partly clouded, 2: misty, 3: light Precipitation, 4: heavy precipitation |
| Temperature | Normalized temperature |
| Humidity | Normalized humidity |

It should be remembered that the discriminative property $x_D$ is not used as a feature in the black-box model and would not have been flagged using standard Shapley values for feature importance.

### 4.3 Shapley values for cluster importance supplement explanations based on feature importance

Here we explain a machine learning model which predicts the daily number of rented bikes based on corresponding weather and seasonal information from a real, publicly available dataset. The predictions in this example are made using simple and intuitive models which in principle should be easy to interpret, but we assume that we have no knowledge about the models which are used, and demonstrate that the explanations produced using Shapley values for cluster importance correspond to our intuitive explanations.

The machine learning model is trained on the Bike Sharing dataset (Fanaee-T and Gama 2013), which comprise data from year 2011 and 2012 in a capital bike-share system. The training data comprises data from the first year, and we use the second year for testing. The available explanatory variables include weather and seasonal information. For simplicity, we concentrate on a selection of the available explanatory variables, and use the five features listed in Table 3.

The training data is illustrated in Fig. 7. Predictions are produced for the points in the test dataset, and we assume that we are asked to explain the predicted count of rented bikes on four days in the test dataset (year 2): day 46, 137, 228 and 320.

#### 4.3.1 Feature importance

Before we explain the predictions using Shapley values for training data cluster importance, we calculate and analyse the Shapley values for feature importance of the four selected days. We use the *iml*-package (Molnar et al 2018) in R (R Core Team 2019), which computes the Shapley values for feature importance following the methodology by Štrumbelj and Kononenko (2014) as described in Sect. 2.3. The results show that both season and temperature significantly affect the predictions. For the first and last explained day (day 46 and day 320), the temperature feature contribute to decrease the predicted number of bike rentals, relative to the mean, while for the two middle days (day 137 and day 228), the temperature feature contribute the most to increase the predicted number of rented bikes.

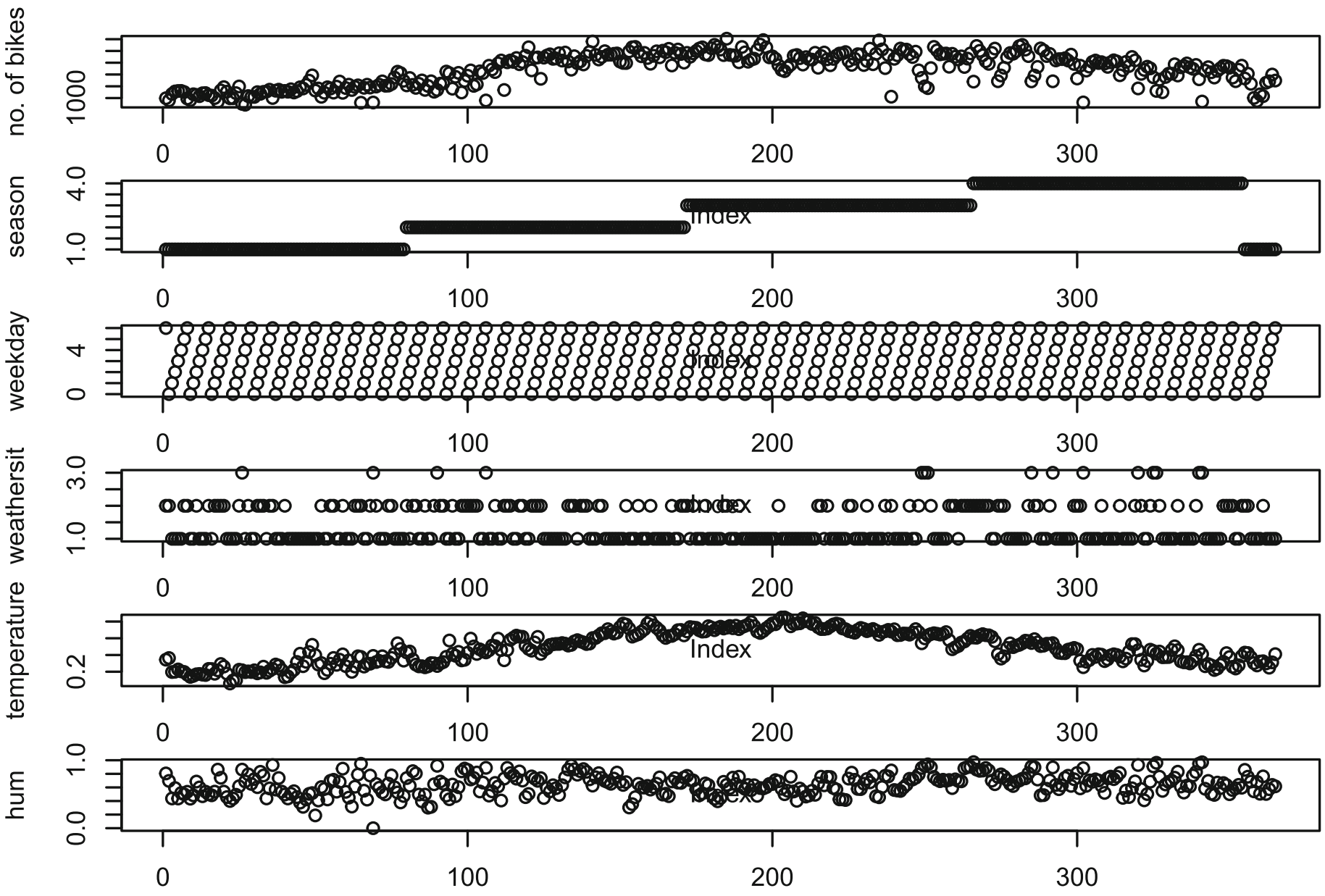

**Fig. 7** Training dataset used in the bike rental example

### 4.3.2 Training data cluster importance

To approach a deeper understanding of how temperature affects the predictions, we propose to calculate and analyse Shapley values for training data cluster importance, and base the clusters on increasing temperature. We choose to use seven different equally sized clusters, ordered by increasing temperature. Clusters or subsets of training data might be created in many different ways, but in this demonstration, we focus on temperature clusters. The clusters of the training data are illustrated in Fig. 8a and b.

The Shapley values for training data cluster importance for the four days of interest are presented in Fig. 9. Here we define the non-distributed gain, $\varphi_0$, to be equal to the mean of the response of the training data. Hence, the Shapley values show how the seven different clusters change the predicted number of rented bikes relative to the mean response in the training data. The upper plot shows the predictions for all days in the test dataset which comprises data from year two. The temperature (normalised) is shown in the second row. The values for the four selected days are marked with red points. The Shapley values for cluster importance are shown in the third row, in ascending order (cluster 1 at bottom (light blue), and cluster 7 at top (orange)). The plots in the bottom row, show how the Shapley value estimates develop when the number of Monte Carlo iterations $m$ is growing from 1 to 250.

**Fig. 8** Training dataset which is clustered based on temperature. The data points' membership in the different clusters are indicated with different colors. **a** Shows trace plots of each feature. Additionally, the response (the number of rented bikes) is shown on top. Note that the observations are sorted according to temperature, hence the numbers on the horizontal axis do not correspond to the days of the year. In **b**, the same data is illustrated with a scatterplotmatrix

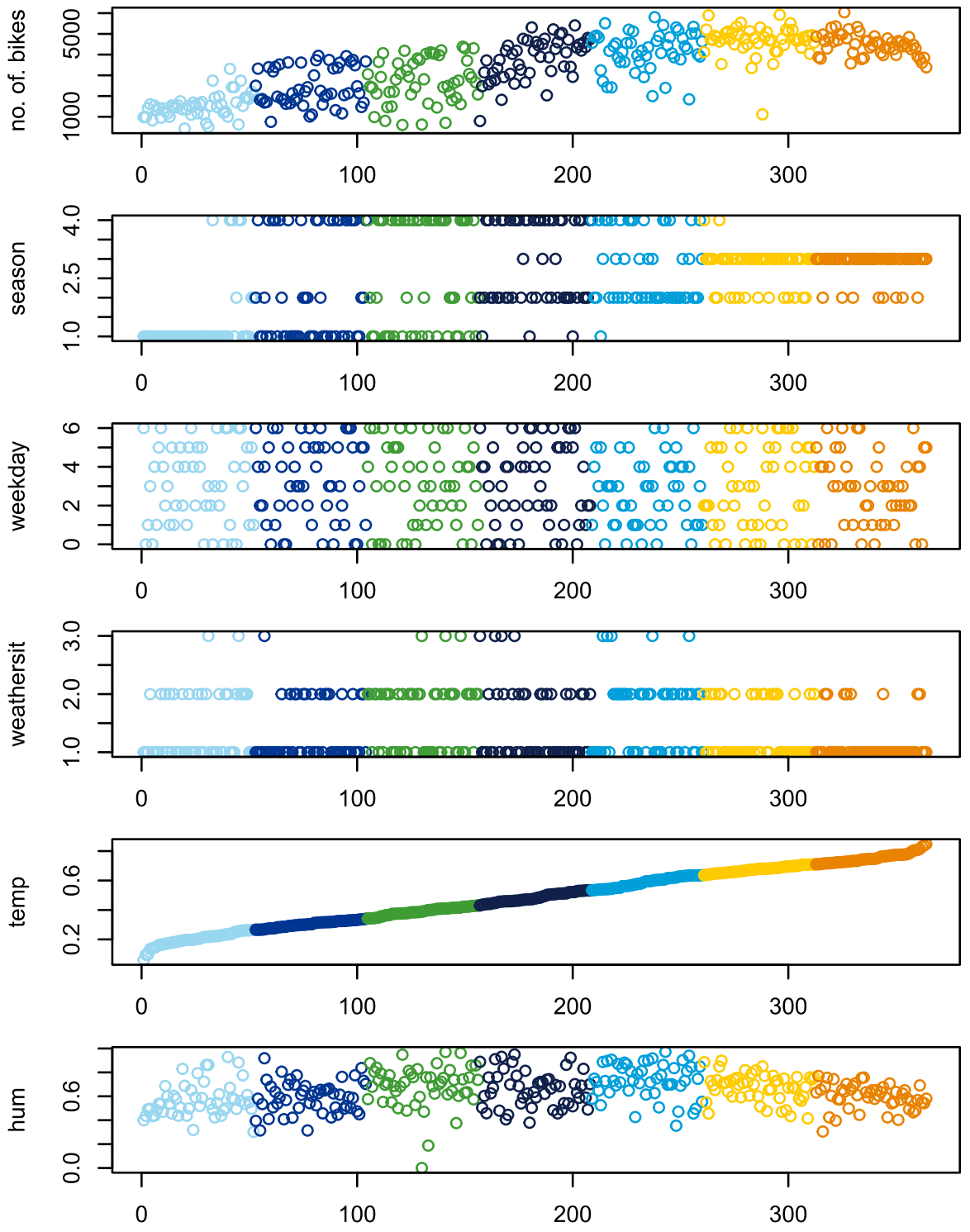

**(a)**

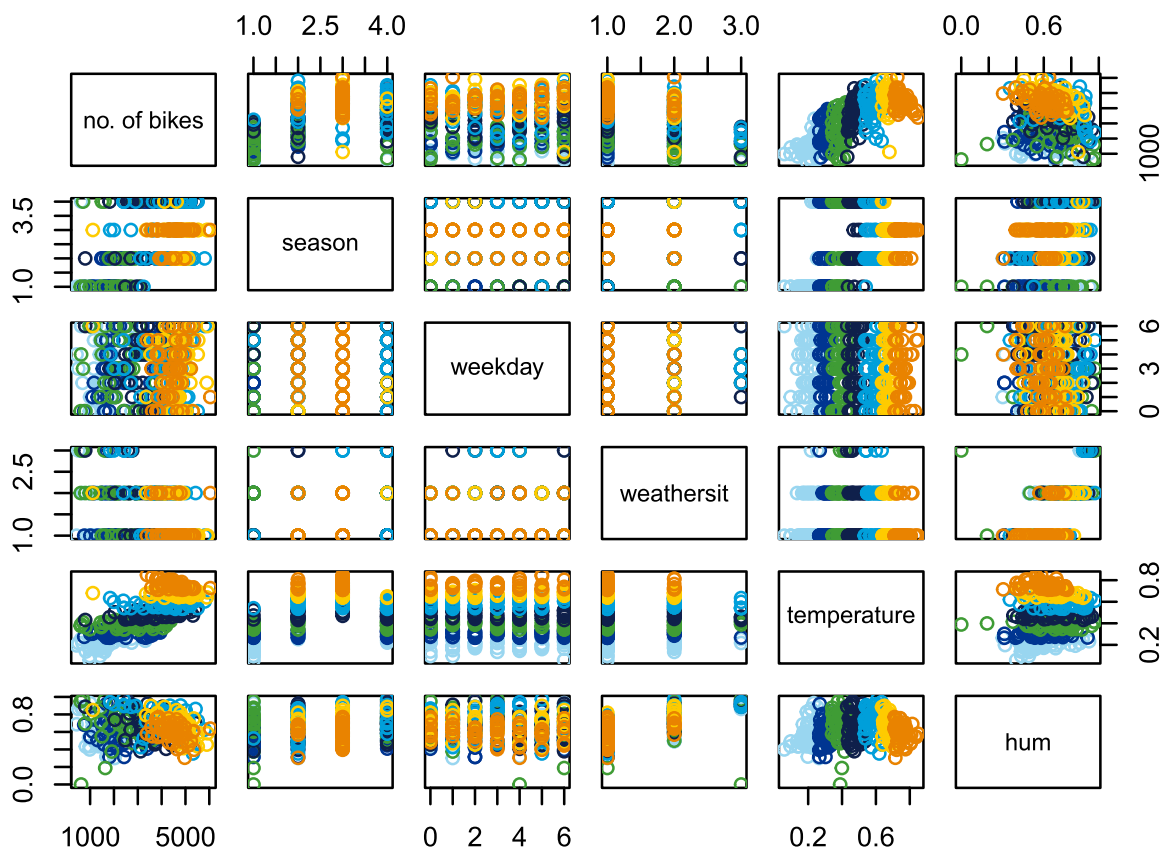

**(b)**

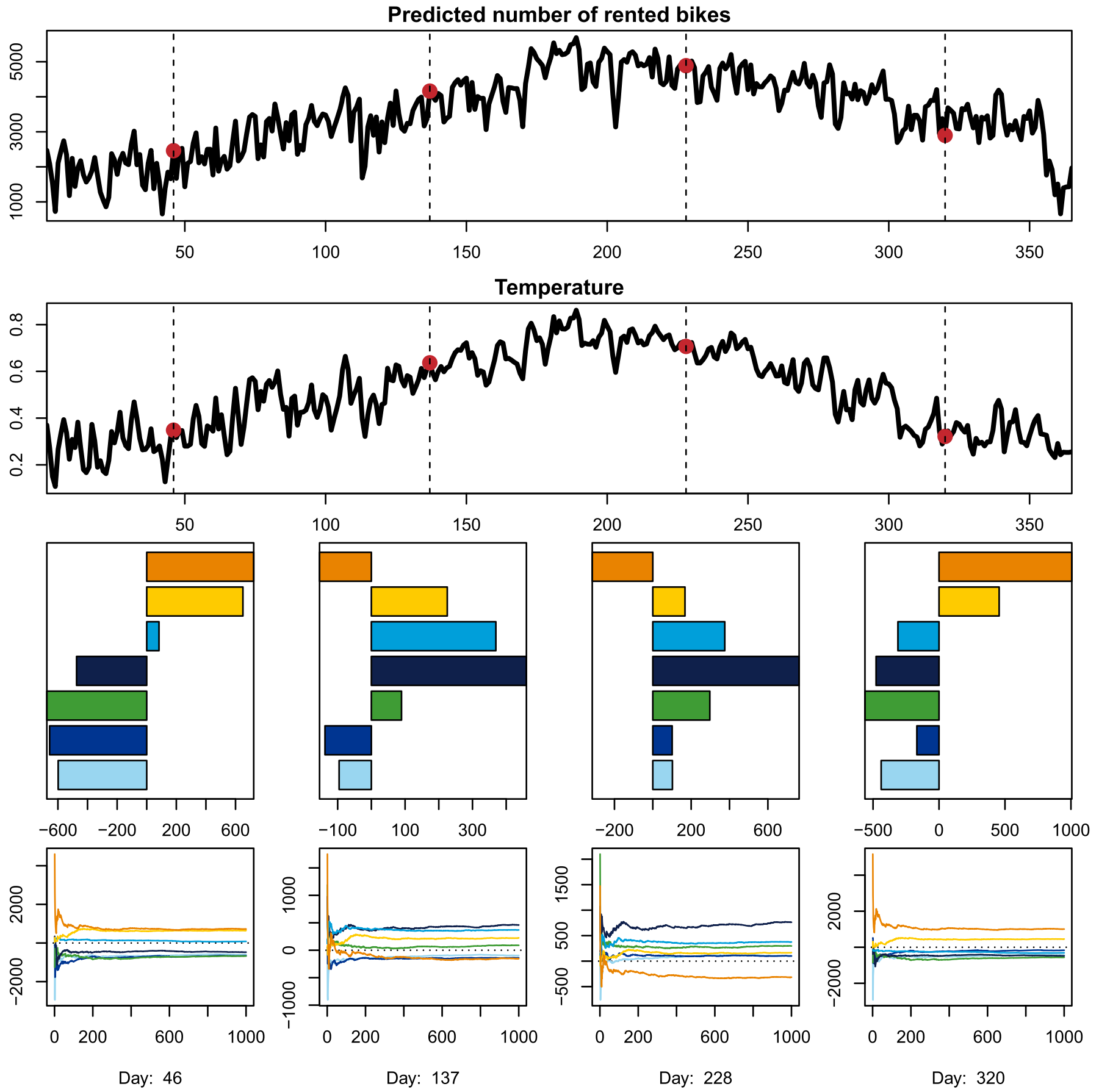

**Fig. 9** Shapley values for training data cluster importance used to explain predictions of a linear regression model on the test dataset (2012). The Shapley values show how the different clusters contribute to change the prediction relative to the mean of the response in training data, $\bar{y} = 3405.762$

In Fig. 9, we observe, for the prediction at day 46, that the clusters which comprise the data with highest temperature (cluster 6 and 7) contribute significantly to increase the prediction. The same applies to the prediction at day 320. Note that the observed temperature is quite low at these days. The predictions at the two middle days (day 137 and 228), however, are not specifically increased due to the training data clusters with the highest temperatures, even though the temperature at the selected days is high. To make it easier for us to assess and evaluate the quality of our explanations, the black-box model used here, is a linear model. Knowing this, and also having a second look at the training data in Fig. 8b, we can argue that the explanations above are reasonable. When fitting a linear model, the slope of the model is not necessarily increased by adding training data instances with high response values. For example, the instances in the seventh cluster, have both high response values and high temperatures, but if we investigate Fig. 8b closely, this cluster seems to decrease the slope. Decreasing the slope, leads to lower predictions for instances with high temperature, while when the temperature is low, the predictions will be higher.

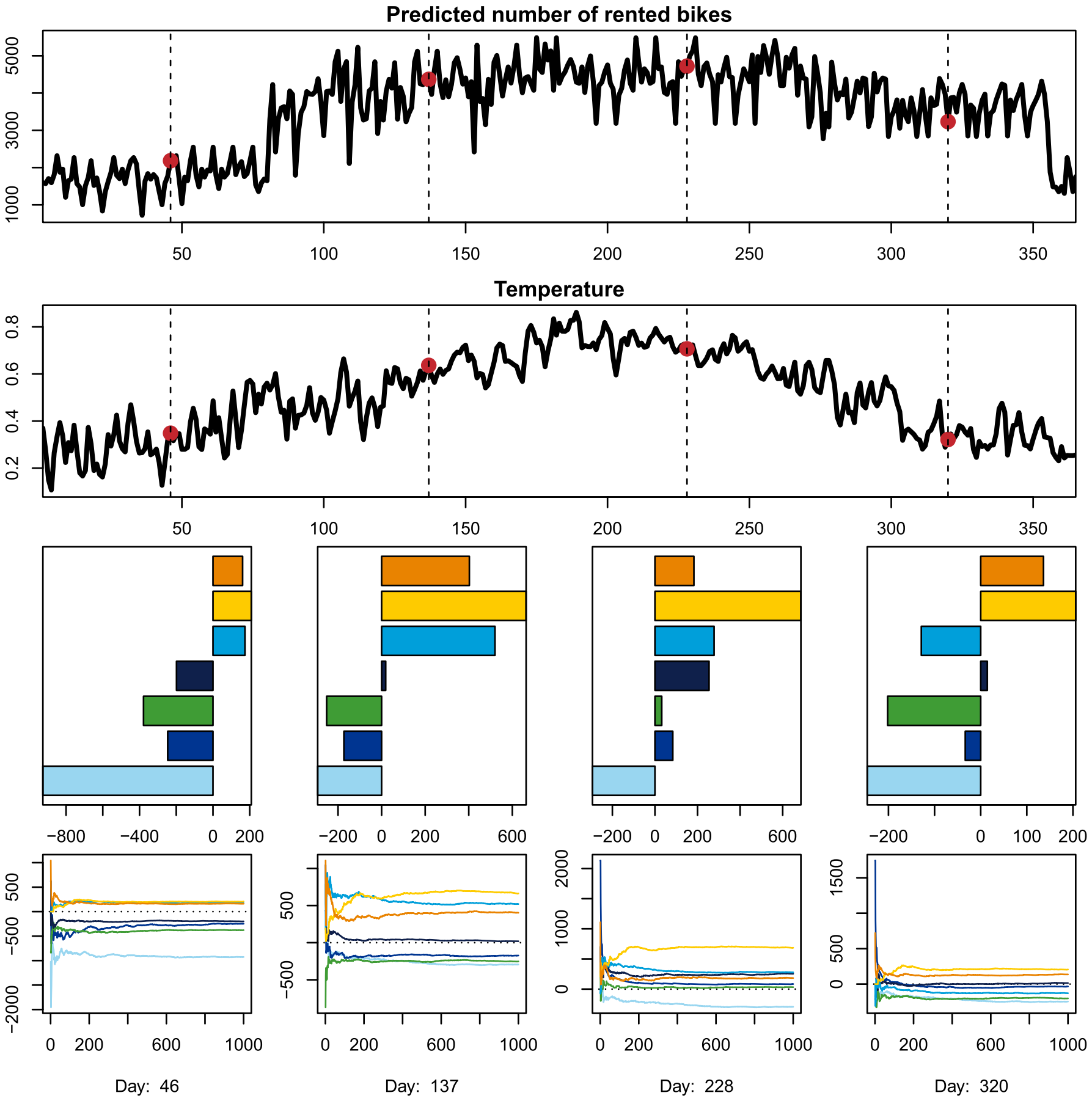

**Fig. 10** Shapley values for training data cluster importance used to explain predictions of a nearest neighbor model with 3 neighbors. The Shapley values show how the different clusters contribute to change the prediction relative to the mean of the response in training data, $\bar{y} = 3405.762$

Suppose now that the linear model is replaced by a nearest neighbor model. The predictions and corresponding explanations are shown in Fig. 10. Changing the model, obviously leads to different predictions and explanations. Now, the clusters with high temperature contributes to high predictions, and the clusters with low temperature contributes to low predictions.

Obviously, linear models and nearest neighbor models are in principle easy to interpret, and one can argue that such models do not need any further explanations. Remember that in reality we do not know which models are used and treat the models as black-boxes. The reason for choosing to explain these simple models is to demonstrate that the explanations using Shapley values for training data cluster importance correspond to the intuitive explanations.

## 5 Discussion and extensions

The presented Shapley values for training data cluster importance satisfies a set of reasonable properties, and we demonstrate that the explanations are as expected on a set of simplistic examples. However, verifying the correctness or soundness of the Shapley values for cluster importance is challenging, especially for larger real-world applications. This is the case for any explanations or interpretation technique, and no single set of evaluation metrics can be applied to all explanation methods (Zhou et al 2021). Future work should include implementation of more real-world applications and experiments. When applicable, human subject evaluation should be performed to evaluate to what extent humans, both experts and lay users, can make use of the Shapley values for cluster importance in practice to increase their understanding and insight about the black-box model. In cases where data-driven models are used to provide decision support (to human decision makers), it can also be possible to evaluate the quality of the explanations by investigating if humans who receive an explanation make better decisions.

In addition, a set of extensions could be explored, some of which are presented below.

### 5.1 Combined shapley values for feature and cluster importance

It is possible to construct a combined Shapley value for training data cluster and feature importance. To evaluate the importance of a feature $j$ of a regression function $f : \mathcal{A} \to \mathbb{R}$, and at the same time, the importance of the training data in a cluster $k$, we define a value function $v(S, W)$ which is the expectation of $f$ when it has seen $x \in \mathcal{A}$ for the features in subset $S \subseteq \{\mathcal{A}_1, \ldots, \mathcal{A}_J\}$, and $f$ is trained on a dataset composed by the union of clusters $Q_k$ for $k \in S \subseteq \{1, \ldots, K\}$.

We define the Shapley value of feature $j$ and cluster $k$ by combining Eq. (8) and Eq. (11),

$$\varphi_{jk}(x) = \frac{1}{K!}\frac{1}{J!}\sum_{\mathcal{B}\in\pi(K)}\sum_{\mathcal{O}\in\pi(J)}\sum_{z\in\mathcal{A}} p(z)\cdot\Big[f_{\text{Pre}^k(\mathcal{B})\cup\{k\})}(\tau(x, z, \text{Pre}^j(\mathcal{O})\cup\{j\})) - f_{\text{Pre}^k(\mathcal{B}))}(\tau(x, z, \text{Pre}^j(\mathcal{O})))\Big], \tag{17}$$

where $\pi(K)$ is the set of all permutations of $K$ clusters, and $\text{Pre}^k(\mathcal{O})$ is the set of all clusters which precede the $k$-th cluster in permutation $\mathcal{O} \in \pi(K)$. Furthermore, $\pi(J)$ is the set of all permutations of the $J$ different features, and $\text{Pre}^j(\mathcal{O})$ is the set of all features which precede the $j$-th feature in permutation $\mathcal{O} \in \pi(J)$. Approximation of Eq. (17) can be accomplished with simulations following the procedure we described in Sect. 2.3 to approximate the Shapley value for feature importance. A further study of the combined Shapley value for feature and cluster importance, including interpretation and application, should be a topic for future work.

### 5.2 Alternative formulation using random values

In this paper, we calculate the Shapley values for cluster importance by comparing the predictions $f_S$ of a function which is trained on a subset $S$ of the available training data, with the predictions $f_{S\cup\{k\}}$, which is trained on a dataset which in addition comprise the data of subset $k$. This approach is in line with existing work on influence functions (Koh and Liang 2017). An alternative formulation is to let $f_S$ be trained on a dataset which consists of the full training dataset, but where the rows which correspond to the data points not comprised in a subset $k \in S$ are replaced by random values inspired by the traditional approach to calculate Shapley values for feature importance. A practical issue arises concerning how to sample both response and features randomly. We have implemented and investigated one version of this alternative formulation, and in the examples we have encountered, the two approaches produce similar explanations. We encourage further investigation of this.

### 5.3 Extended learning curves

Two types of learning curves appears in literature. The first type visualizes the performance of an iterative machine learning algorithm as a function of its training time or number of iterations. The second type, which we concentrate on, is used to extrapolate performance from smaller to larger datasets (Domhan et al 2015). Usually, the number of samples are shown on the horizontal axis, and the vertical axis shows a metric for the predictive power, for example mean squared error. Patterns which depend on the size of the training dataset are sometimes evident across different datasets, and such patterns can be discovered through learning curve analysis (Perlich et al 2003; Kolachina et al 2012). When learning curves are drawn, the underlying training data is often grown only once. However, growing the dataset in a different way, will sometimes significantly change the shape of the learning curve. This information is typically not conveyed by traditional learning curves. When we calculate our Shapley values, the model is retrained $M$ times, using different training datasets of different size. We can plot this information in a scatter plot, similar to a traditional learning curve, with size of training data on the horizontal axis, and the performance metric on the vertical axis. By doing this, more information about the data and the algorithm's learning process can be disclosed to the user, which can enable more informed and possibly more accurate decisions.

### 5.4 Efficiency

The proposed approach for approximating the Shapley values is computationally expensive, as the model is retrained for each sample $m \in \{1, \dots, M\}$. Hence, future work should consider more efficient procedures to approximate the Shapley values. This can include utilising the property that the models are trained on coalitions of clusters which are order independent. Also, instead of re-initialising the model for each sample, procedures to reuse the weights from a previous sample should be explored.

This can perhaps be relevant for at least some optimisation methods, like gradient-based methods that incrementally update the weights.

### 5.5 Clusters of different sizes

In this paper, we only consider Shapley values for cluster importance for equisized clusters. Clusters of different sizes can also be considered, and calculated following the same procedure as for equisized clusters. The user should, however, take the cluster size into consideration when interpreting the results. Furthermore, Shapley values are not always robust against merging and splitting (Knudsen and Østerdal 2012), meaning that the sum of the Shapley value of cluster $i$ and $j$ can differ (slightly) from the Shapley value of the merged cluster $k = i \cup j$, that is $\varphi_i + \varphi_j \neq \varphi_k$.

## 6 Conclusion

In this paper, we have proposed a novel model-agnostic methodology to explain individual predictions from black-box machine learning models. The proposed methodology quantifies how different clusters in the training data affect individual predictions. A set of examples are presented to illustrate and explain the methodology, demonstrating that predictions of data with a known signal generating function are accurately explained. We have presented examples with simple and transparent models which we intuitively understand, and shown that the explanations provided by the Shapley values for cluster importance correspond to these intuitive explanations. Furthermore, that Shapley values for cluster importance can be used to reveal biased behavior and erroneous training data. The novel approach proposed in this paper allows us to explore and investigate how the training data affects the predictions made by any black-box model. New aspects of the reasoning and inner workings of a prediction model and learning method can be conveyed. This is insight which would not be available without the proposed methodology, and should complement existing explanations offered by measures of feature importance.

**Acknowledgements** The authors greatly appreciate the suggestions and criticisms by Odin Gramstad (DNV), Martin Jullum (Norwegian Computing Center), Arnoldo Frigessi (University of Oslo), Mette Langaas (Norwegian University of Science and Technology), Knut Erik Knutsen (DNV) and Erik Vanem (DNV/University of Oslo). The work is carried out in collaboration with the Big Insight SFI, and is partly funded by the Research Council of Norway (project number 237718 and 251396) and DNV.

**Funding** Open access funding provided by Volda University College

## Appendix A: Exact solution to the example presented in Sect. 4.1.1 using a linear model

See Tables 4, 5, 6.

**Table 4** *Linear model, Cluster 1:* Calculation of the exact Shapley value for cluster importance of cluster 1: $\varphi_1 = 1/6(11.95{+}11.95{+}3.63{+}2.12 + 3.94 + 2.12) = 5.95$

| $\mathcal{O}$ | $S \cup \{k\}$ | $S$ | $f_{\mathcal{O}\cup\{k\}}$ | $f_{\mathcal{O}}$ | $f_{\mathcal{O}\cup\{k\}} - f_{\mathcal{O}}$ |
|---|---|---|---|---|---|
| {1 2 3} | {1, 2} | {1} | 11.95 | 0.00 | 11.95 |
| {1 3 2} | {1, 2, 3} | {1, 3} | 11.95 | 0.00 | 11.95 |
| {2 1 3} | {2} | Ø | 9.61 | 5.98 | 3.63 |
| {2 3 1} | {2} | Ø | 7.72 | 5.60 | 2.12 |
| {3 1 2} | {1, 2, 3} | {1, 3} | 8.45 | 4.51 | 3.94 |
| {3 2 1} | {2, 3} | {3} | 7.72 | 5.60 | 2.12 |

**Table 5** *Linear model, Cluster 2:* Calculation of the exact Shapley value for cluster importance of cluster 2: $\varphi_2 = 1/6(-2.34 + -0.74 + 5.98 + 5.98 + -0.74 + 1.08) = 1.54$

| $\mathcal{O}$ | $S \cup \{k\}$ | $S$ | $f_{\mathcal{O}\cup\{k\}}$ | $f_{\mathcal{O}}$ | $f_{\mathcal{O}\cup\{k\}} - f_{\mathcal{O}}$ |
|---|---|---|---|---|---|
| {1 2 3} | {1, 2} | {1} | 9.61 | 11.95 | −2.34 |
| {1 3 2} | {1, 2, 3} | {1, 3} | 7.72 | 8.45 | −0.74 |
| {2 1 3} | {2} | Ø | 5.98 | 0 | 5.98 |
| {2 3 1} | {2} | Ø | 5.98 | 0 | 5.98 |
| {3 1 2} | {1, 2, 3} | {1, 3} | 7.72 | 8.45 | −0.74 |
| {3 2 1} | {2, 3} | {3} | 5.60 | 4.51 | 1.08 |

**Table 6** *Linear model, Cluster 3:* Calculation of the exact Shapley value for cluster importance of cluster 3: $\varphi_3 = 1/6(-1.90 + -3.50 + -1.90 + -0.38 + 4.51 + 4.51) = 0.23$

| $\mathcal{O}$ | $S \cup \{k\}$ | $S$ | $f_{\mathcal{O}\cup\{k\}}$ | $f_{\mathcal{O}}$ | $f_{\mathcal{O}\cup\{k\}} - f_{\mathcal{O}}$ |
|---|---|---|---|---|---|
| {1 2 3} | {1, 2, 3} | {1, 2} | 7.72 | 9.61 | −1.90 |
| {1 3 2} | {1, 3} | {1} | 8.45 | 11.95 | −3.50 |
| {2 1 3} | {1, 2, 3} | {1, 2} | 7.72 | 9.61 | −1.90 |
| {2 3 1} | {2, 3} | {2} | 5.60 | 5.98 | −0.38 |
| {3 1 2} | {3} | Ø | 4.51 | 0 | 4.51 |
| {3 2 1} | {3} | Ø | 4.51 | 0 | 4.51 |

## Appendix B: Exact solution to the example presented in Sect. 4.1.1 using a *k*NN model

See Tables 7, 8, 9.

**Table 7** *kNN, Cluster* 1: Calculation of the exact Shapley value for cluster importance of cluster 1: $\varphi_1 = 1/6(10.24+10.24+0.00+0.00+0.00+0.00) = 3.41$

| $\mathcal{O}$ | $S \cup \{k\}$ | $S$ | $f_{\mathcal{O}\cup\{k\}}$ | $f_{\mathcal{O}}$ | $f_{\mathcal{O}\cup\{k\}} - f_{\mathcal{O}}$ |
|---|---|---|---|---|---|
| {1 2 3} | {1, 2} | {1} | 10.24 | 0.00 | 10.24 |
| {1 3 2} | {1, 2, 3} | {1, 3} | 10.24 | 0.00 | 10.24 |
| {2 1 3} | {2} | Ø | 8.15 | 8.15 | 0.00 |
| {2 3 1} | {2} | Ø | 4.71 | 4.71 | 0.00 |
| {3 1 2} | {1, 2, 3} | {1, 3} | 4.71 | 4.71 | 0.00 |
| {3 2 1} | {2, 3} | {3} | 4.71 | 4.71 | 0.00 |

**Table 8** *kNN, Cluster* 2: Calculation of the exact Shapley value for cluster importance of cluster 2: $\varphi_2 = 1/6(-2.09+0.00+8.15+8.15+0.00+0.00) = 2.37$

| $\mathcal{O}$ | $S \cup \{k\}$ | $S$ | $f_{\mathcal{O}\cup\{k\}}$ | $f_{\mathcal{O}}$ | $f_{\mathcal{O}\cup\{k\}} - f_{\mathcal{O}}$ |
|---|---|---|---|---|---|
| {1 2 3} | {1, 2} | {1} | 8.15 | 10.24 | −2.09 |
| {1 3 2} | {1, 2, 3} | {1, 3} | 4.71 | 4.71 | 0.00 |
| {2 1 3} | {2} | Ø | 8.15 | 0.00 | 8.15 |
| {2 3 1} | {2} | Ø | 8.15 | 0.00 | 8.15 |
| {3 1 2} | {1, 2, 3} | {1, 3} | 4.71 | 4.71 | 0.00 |
| {3 2 1} | {2, 3} | {3} | 4.71 | 4.71 | 0.00 |

**Table 9** *kNN, Cluster* 3: Calculation of the exact Shapley value for cluster importance of cluster 3: $\varphi_3 = 1/6(-3.44-5.53-3.44-3.44+4.71+4.71) = -1.07$

| $\mathcal{O}$ | $S \cup \{k\}$ | $S$ | $f_{\mathcal{O}\cup\{k\}}$ | $f_{\mathcal{O}}$ | $f_{\mathcal{O}\cup\{k\}} - f_{\mathcal{O}}$ |
|---|---|---|---|---|---|
| {1 2 3} | {1, 2} | {1} | 4.71 | 8.15 | −3.44 |
| {1 3 2} | {1, 2, 3} | {1, 3} | 4.71 | 10.24 | −5.53 |
| {2 1 3} | {2} | Ø | 4.71 | 8.15 | −3.44 |
| {2 3 1} | {2} | Ø | 4.71 | 8.15 | −3.44 |
| {3 1 2} | {1, 2, 3} | {1, 3} | 4.71 | 0.00 | 4.71 |
| {3 2 1} | {2, 3} | {3} | 4.71 | 0.00 | 4.71 |

## Appendix C: Dataset for the example presented in Sect. 4.1

See Table 10.

**Table 10** Dataset

| x | y |
|---|---|
| 0.77 | 7.86 |
| 1.99 | 9.75 |
| 2.86 | 10.24 |
| 5.49 | 14.74 |
| 6.48 | 15.72 |
| 6.75 | 13.79 |
| 0.50 | 1.85 |
| 1.87 | 4.31 |
| 2.55 | 4.01 |
| 5.08 | 8.15 |
| 6.41 | 10.45 |
| 7.28 | 10.95 |
| 0.59 | −0.14 |
| 2.44 | 4.26 |
| 3.30 | 4.71 |
| 5.38 | 7.04 |
| 5.72 | 5.38 |
| 6.59 | 5.88 |

**Publisher's Note** Springer Nature remains neutral with regard to jurisdictional claims in published maps and institutional affiliations.